\pdfoutput=1

\documentclass[11pt]{article}

\usepackage{ACL2023}

\usepackage{times}
\usepackage{latexsym}
\usepackage{multirow}
\usepackage{makecell}

\usepackage[T1]{fontenc}

\usepackage[utf8]{inputenc}

\usepackage{microtype}

\usepackage{inconsolata}
\usepackage{graphicx}

\newcommand{\ie}{\textit{i}.\textit{e}.,\ }
\newcommand{\eg}{\textit{e}.\textit{g}.,\ }

\title{LayoutMask: Enhance Text-Layout Interaction in Multi-modal Pre-training for Document Understanding}

\author{Yi Tu, Ya Guo, Huan Chen, Jinyang Tang\\
  Ant Group, China \\
  \texttt{\{qianyi.ty,guoya.gy,chenhuan.chen,jinyang.tjy\}@antgroup.com} 
  }

\begin{document}

\maketitle
\begin{abstract}
Visually-rich Document Understanding (VrDU) has attracted much research attention over the past years.
Pre-trained models on a large number of document images with transformer-based backbones have led to significant performance gains in this field.
The major challenge is how to fusion the different modalities (text, layout, and image) of the documents in a unified model with different pre-training tasks. 
This paper focuses on improving text-layout interactions and proposes a novel multi-modal pre-training model, LayoutMask.
LayoutMask uses local 1D position, instead of global 1D position, as layout input and has two pre-training objectives: (1) Masked Language Modeling: predicting masked tokens with two novel masking strategies; (2) Masked Position Modeling: predicting masked 2D positions to improve layout representation learning.
LayoutMask can enhance the interactions between text and layout modalities in a unified model and produce adaptive and robust multi-modal representations for downstream tasks.
Experimental results show that our proposed method can achieve state-of-the-art results on a wide variety of VrDU problems, including form understanding, receipt understanding, and document image classification.
\end{abstract}

\section{Introduction}
\label{sec:intro}

\begin{figure}[t]
	\centering
	\includegraphics[width=7.5cm]{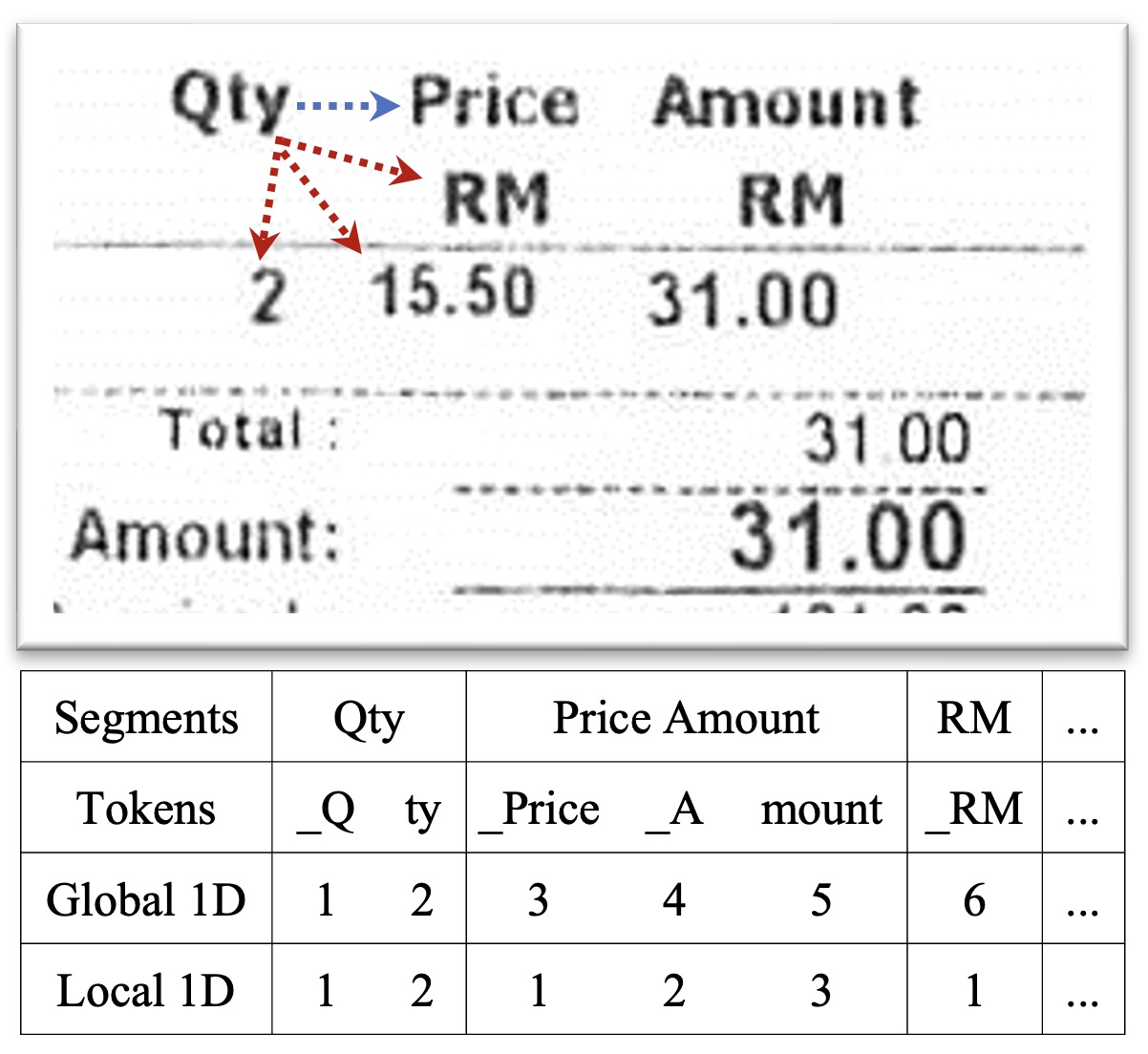}
	\caption{A receipt image from SROIE dataset and the global/local 1D positions of tokens based on global/in-segment reading orders. Local 1D positions restart with ``1'' for each individual segment.
	\textbf{Blue Arrow:} When using global 1D position, the  reading order is explicitly implied by the ascending numbers, so the word after ``Qty'' is ``Price''.
	\textbf{Red Arrows:} When using local 1D position, the successor of ``Qty'' is not directly given and can have more possible  choices, so their semantic relations and 2D positions will be considered during pre-training.}
	\label{fig:global-vs-local}
\end{figure}

Visually-rich Document Understanding (VrDU) is an important research area that aims to understand various types of documents (\eg forms, receipts, and posters), and it has attracted much attention from both academia and industry. 
In recent years, pre-training techniques \citep{kenton2019bert, zhang2019ernie} have been introduced into this area and
self-supervised pre-training multi-modal models have demonstrated great successes in various VrDU tasks \citep{xu2020layoutlm, xu2021layoutlmv2,hong2022bros,li2021structurallm}.

However, existing document pre-training models suffer from reading order issues.
 Following the idea of BERT \citep{kenton2019bert}, these methods \citep{xu2020layoutlm, xu2021layoutlmv2, hong2022bros} usually adopt ascending numbers (\eg 0, 1, 2,.., 511) to represent the global reading order of tokens in the document.
Then, these numbers are encoded into 1D position embeddings to provide explicit reading order supervision during pre-training, which are called ``global 1D position''.
While such global 1D positions are widely used in NLP models for textual data, it is not a good choice for document data. 
Firstly, plain texts always have definite and linear reading orders, but the reading order of a document may not be unique or even linear, which cannot be simply encoded with monotonically increasing numbers.
Secondly, the global reading order of a document is usually obtained by ordering detected text segments from OCR tools with empirical rules, so it heavily relies on stable and consistent OCR results, affecting the generalization ability in real-world applications.
Moreover, the empirical rules to obtain reading orders (\eg ``top-down and left-right'') may not be able to handle documents with complex layouts, thus providing inaccurate supervision. 

Some previous studies have attempted to solve the above reading order issues.
LayoutReader \citep{wang2021layoutreader} proposes a sequence-to-sequence framework for reading order detection with supervised reading order annotations.
XYLayoutLM \citep{gu2022xylayoutlm} utilizes an augmented XY Cut algorithm to generate different proper reading orders during pre-training to increase generalization ability.
ERNIE-Layout \citep{peng2022ernie} rearranges the order of input tokens in serialization modules and adopts a reading order prediction task in pre-training.
While these studies propose data-based or rule-based solutions to provide explicit reading order supervision, we believe that the self-supervised pre-training process on a large number of documents without using extra supervision is sufficient to help the model to learn reading order knowledge, and such knowledge can be implicitly encoded into the pre-trained model with better adaptiveness and robustness to various document layouts.  

We proposed a novel multi-modal pre-training model, \textbf{LayoutMask}, to achieve this goal.
LayoutMask only uses text and layout information as model input and aims to enhance text-layout interactions and layout representation learning during pre-training.
It differs from previous studies in three aspects: choice of 1D position, masking strategy, and pre-training objective.

Instead of global 1D position, LayoutMask proposes to use the in-segment token orders as 1D position, which is referred to as ``\textbf{local 1D position}'' (See illustration in Figure \ref{fig:global-vs-local}). 
As local 1D position does not provide cross-segment orders, LayoutMask is supposed to infer global reading order by jointly using 1D position, 2D position, and semantic information, thus bringing in-depth text-layout interactions. 
To further promote such interactions, we equip the commonly used pre-training objective, Masked Language Modeling (MLM), with two novel masking strategies, \textbf{Whole Word Masking} and \textbf{Layout-Aware Masking}, and design an auxiliary pre-training objective, \textbf{Masked Position Modeling}, to predict masked 2D positions during pre-training.
With the above designs, we increase the difficulty of pre-training objectives and force the model to focus more on layout information to obtain reading order clues in various document layouts in self-supervised learning, thus producing more adaptive and robust text-layout representations for document understanding tasks.

Experimental results show that our proposed method can bring significant improvements to VrDU tasks and achieve SOTA performance with only text and layout modalities, indicating that previous studies have not fully explored the potential power of layout information and text-layout interactions.
The contributions of this paper are summarized as follows:

\begin{enumerate}
\item We propose LayoutMask, a novel multi-modal pre-training model focusing on text-layout modality, to generate adaptive and robust multi-modal representations for VrDU tasks.
\item In LayoutMask, we use local 1D position instead of global 1D position to promote reading order learning. We leverage Whole Word Masking and Layout-Aware Masking in the MLM task and design a new pre-training objective, Masked Position Modeling, to enhance text-layout interactions.  
\item Our method can produce useful multi-modal representations for documents and significantly outperforms many SOTA methods in multiple VrDU tasks. 

\end{enumerate}

\section{Related Work}
\label{sec:Related Work}
The early studies in VrDU area usually use uni-modal models or multi-modal models with shallow fusion \citep{yang2016hierarchical,yang2017learning,katti2018chargrid,sarkhel2019deterministic}.
In recent years, pre-training techniques in NLP \citep{kenton2019bert, zhang2019ernie, bao2020unilmv2} and CV \citep{bao2021beit,li2022dit} have become more and more popular, and they have been introduced into this area.
Inspired by BERT \citep{kenton2019bert}, LayoutLM \citep{xu2020layoutlm} first improved the masked language modeling task by using the 2D coordinates of each token as layout embeddings, which can jointly model interactions between text and layout information and benefits document understanding tasks.
Following this idea, LayoutLMv2 \citep{xu2021layoutlmv2} propose to concatenate image patches with textual tokens to enhance text-image interactions, and LayoutLMv3 \citep{huang2022layoutlmv3}  proposed to learn cross-modal alignment with unified text and image masking.

While the above methods focus on text-image interactions, some other studies have realized the importance of layout information.
StructuralLM \citep{li2021structurallm} utilizes segment-level layout features to provide word-segment relations.
DocFormer \citep{appalaraju2021docformer} combines text, vision, and spatial features with a novel multi-modal self-attention layer and shares learned spatial embeddings across modalities.
LiLT \citep{wang2022lilt} proposes a language-independent layout transformer where the text and layout information are separately embedded.
ERNIE-Layout \citep{peng2022ernie} adopts a reading order prediction task in pre-training and rearranges the token sequence with the layout knowledge.

\section{Methodology}

\begin{figure*}[ht]
	\centering
	\includegraphics[width=15.5cm]{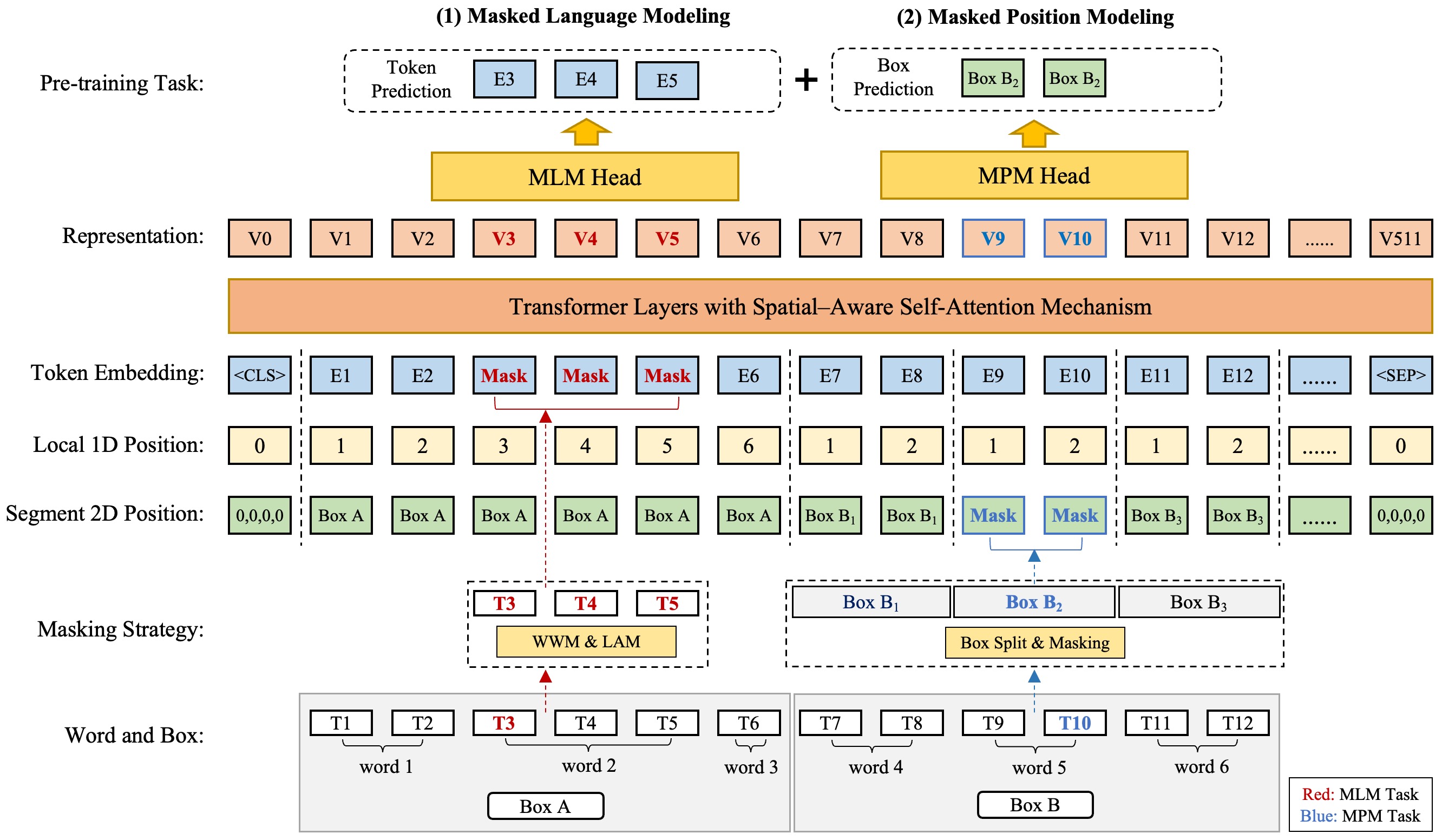}
	\caption{The model pipeline of LayoutMask. \textbf{Red Text:} Illustration of the Masked Language Modeling task. \textbf{Blue Text:} Illustration of the Masked Position Modeling task.}
	\label{fig:model-structure}
\end{figure*}

LayoutMask is a multi-modal transformer that can encode text and layout information of documents and produce multi-modal representations.
The pipeline of LayoutMask can be seen in Figure \ref{fig:model-structure}.
LayoutMask uses the transformer model with a spatial-aware self-attention mechanism proposed in LayoutLmv2 \citep{xu2021layoutlmv2} as the backbone and follows its preprocessing settings for text and layout embeddings.  
In Section \ref{sec:selection-layout}, we will discuss the different choices of layout information in LayoutMask.
In Section \ref{sec:pretrain-tasks}, we will introduce the pre-training tasks and masking strategies used in LayoutMask.

\subsection{Selection of Layout Information}
\label{sec:selection-layout}

\begin{table}[]
\footnotesize
\begin{tabular}{l|cc}
\hline
\multirow{2}{*}{\textbf{Method} } & \multicolumn{2}{c}{\textbf{Position}} \\ \cline{2-3}
     & \textbf{1D} & \textbf{2D} \\ \hline

LayoutLM  \citep{xu2020layoutlm}        & Global          & Word         \\

StructuralLM \citep{li2021structurallm}       & Global        & Segment       \\
LayoutLMv2  \citep{xu2021layoutlmv2}      & Global        & Word         \\
BROS   \citep{hong2022bros}        & Global          & Segment$^{\dag}$       \\
LiLT  \citep{wang2022lilt}         & Global        & Segment       \\
LayoutLMv3   \citep{huang2022layoutlmv3}      & Global        & Segment       \\ \hline
\textbf{LayoutMask(Ours)} & \textbf{Local}    & \textbf{Segment}   \\ \hline
\end{tabular}
\caption{\label{table:survey-of-positions}
The 1D position and 2D position choices in previous studies. Our method uses local 1D position and segment-level 2D position. $\dag$: BROS leverages relative 2D positions instead of absolute positions.
}
\end{table}

For VrDU tasks, there are two types of commonly used layout information: 1D position and 2D position. 
We list the 1D and 2D positions used in previous studies in Table \ref{table:survey-of-positions}.

\textbf{1D Position:}
As we discussed in Section \ref{sec:intro}, using global 1D position will bring read order issues and could damage the adaptiveness and robustness of pre-trained models.
Different from some previous models that leverage global 1D position as model input, we propose to use local 1D position in LayoutMask.
Local 1D position only encodes the token orders within each segment and always restarts with 1 for each individual segment.
 Illustrations of the global and local 1D positions can be seen in Figure \ref{fig:global-vs-local} and Figure \ref{fig:model-structure}. 
 Compared with global 1D position, the major difference of using local 1D position is the lack of cross-segment orders, so the global reading order has to be inferred with other layout and semantic clues. Besides, the in-segment orders implied by local 1D position are more reliable and trustworthy than cross-segment orders when meeting complex document layouts.
 
 \textbf{2D Position:}
The 2D position is represented as a 4-digit vector like $[x_1, y_1, x_2, y_2]$, where $[x_1, y_1]$ and $[x_2, y_2]$ are the normalized coordinates of the top-left and bottom-right corners of a text box.
There are two commonly used types of 2D positions: word-level 2D position (Word-2D) and segment-level 2D position (Segment-2D). 
For Word-2D, tokens of the same word will have the same word-level boxes as their 2D position. While for Segment-2D, the segment coordinates are shared by tokens within each segment.   

In our model, we choose local 1D position and segment-level 2D position as our model input, where local 1D position can provide in-segment orders, and segment-level 2D position can provide cross-segment reading order clues, so the pre-trained model can learn the correct global reading order by jointly using 1D and 2D positions.  
We will compare the experimental results using different 1D \& 2D position combinations in Section \ref{sec:layout-ablation} and provide detailed discussions. 

\subsection{Pre-training Objectives}
\label{sec:pretrain-tasks}
\subsubsection{Masked Language Modeling}
\label{sec:mlm}
The Masked Language Modeling task is the most essential and commonly used pre-training task in multi-modal pre-training.
In this task, we randomly mask some tokens with a given probability $\mathrm{P}_\mathrm{mlm}$ (\eg 15\%) and recover these tokens during pre-training.

For each document,  we use $M$ to denote the number of masked tokens.  $\mathrm{y}_i$ and $\mathrm{\bar{y}}_i$ represent the ground truth and prediction of the $i$-th masked token. Then the loss of this task is the average cross entropy loss of all masked tokens:
\begin{equation}
\label{equ:mlm}
\mathcal{L}_\mathrm{mlm}= 
- \frac{1}{M}
 \sum_{i=1}^{M} 
 \mathrm{CE}(\mathrm{y}_i, \mathrm{\bar{y}}_i).
\end{equation}

In preliminary experiments, we find that the naive MLM  task is not optimal for multi-modal pre-training.
Thus we propose to adopt two novel strategies, Whole Word Masking (WWM) and Layout-Aware Masking (LAM), to enhance this task.
  
\noindent\textbf{Whole Word Masking:}
The WWM strategy was first proposed for Chinese language models to increase the task difficulty \citep{cui2021pre}.
Following this strategy,  we set masks at word-level instead of token-level, which is much more challenging.
When using WWM, the semantic relations between masked and unmasked tokens of the same words are eliminated, so the model has to find more context to predict masked words, which can promote text-layout interactions.  

\noindent\textbf{Layout-Aware Masking:}
As we use Local-1D and Segment-2D as model input, the global reading order should be obtained by jointly using 1D and 2D positions, where Local-1D provides in-segment orders and segment-2D provides cross-segment clues.
We find that the cross-segment orders are harder to be learned, so we propose Layout-Aware Masking (LAM) strategy to address this issue.
Unlike naive masking strategy where each token has an equal masking probability $\mathrm{P}_\mathrm{mlm}$,  in LAM strategy, the first and last word of each segment has a higher probability (\ie $3\times\mathrm{P}_\mathrm{mlm}$) to be masked. 
In order to predict such masked words, the model has to pay more attention to finding their contexts in the preceding or succeeding segment, thus promoting learning cross-segment orders.   

\begin{table*}
\small
\centering
\begin{tabular}{l|c|l|ccc}
\hline
\multicolumn{1}{c|}{\textbf{Method}}   & \textbf{\#Parameters} & \textbf{Modality} & \textbf{FUNSD(F1$\uparrow$)} & \textbf{CORD(F1$\uparrow$)} & \textbf{SROIE(F1$\uparrow$)} \\ \hline
BERT$_\mathrm{Base}$   \citep{kenton2019bert}           & 110M                  & T                 & 60.26          & 89.68         & 90.99          \\
RoBERTA$_\mathrm{Base}$ \citep{liu2019roberta}           & 125M                  & T                 & 66.48          & 93.54         & -              \\
UniLMv2$_\mathrm{Base}$    \citep{bao2020unilmv2}       & 125M                  & T                 & 68.90          & 90.92         & 94.59          \\
BROS$_\mathrm{Base}$   \citep{hong2022bros}             & 110M                  & T+L               & 83.05          & 95.73         & 95.48          \\
LiLT$_\mathrm{Base}$    \citep{wang2022lilt}          & -                     & T+L               & 88.41          & 96.07         & -              \\
LayoutLM$_\mathrm{Base}$   \citep{xu2020layoutlm}        & 160M                  & T+L+I             & 79.27          & -             & 94.38          \\
LayoutLMv2$_\mathrm{Base}$ \citep{xu2021layoutlmv2}       & 200M                  & T+L+I             & 82.76          & 94.95         & 96.25          \\
TILT$_\mathrm{Base}$   \citep{powalski2021going}            & 230M                  & T+L+I             &  -              & 95.11         & 97.65$^\dag$          \\
DocFormer$_\mathrm{Base}$ \citep{appalaraju2021docformer}        & 183M                  & T+L+I             & 83.34          & 96.33         & -              \\
LayoutLMv3$_\mathrm{Base}$  \citep{huang2022layoutlmv3}       & 133M                  & T+L+I             & 90.29          & 96.56         & -              \\ \hline
\textbf{LayoutMask$_\mathrm{Base}$ (Ours)}        & 182M                   & T+L               & \textbf{92.91$\pm$0.34}     & \textbf{96.99$\pm$0.30}    & \textbf{96.87$\pm$0.19}     \\ \hline 
BERT$_\mathrm{Large}$    \citep{kenton2019bert}               & 340M                  & T                 & 65.63          & 90.25         & 92.00          \\
RoBERTA$_\mathrm{Large}$ \citep{liu2019roberta}             & 355M                  & T                 & 70.72          & 93.80         & -               \\
UniLMv2$_\mathrm{Large}$     \citep{bao2020unilmv2}          & 355M                  & T                 & 72.57          & 92.05         & 94.88          \\

LayoutLM$_\mathrm{Large}$  \citep{xu2020layoutlm}    & 343M                  & T+L               & 77.89          &-               & 95.24          \\
BROS$_\mathrm{Large}$ \citep{hong2022bros}              & 340M                  & T+L               & 84.52          & 97.40          &-                \\
LayoutLMv2$_\mathrm{Large}$  \citep{xu2021layoutlmv2}        & 426M                  & T+L+I             & 84.2           & 96.01         & \textbf{97.81}          \\
TILT$_\mathrm{Large}$  \citep{powalski2021going}             & 780M                  & T+L+I             & -              & 96.33         & 98.10$^\dag$          \\
DocFormer$_\mathrm{Large}$  \citep{appalaraju2021docformer}        & 536M                  & T+L+I             & 84.55          & 96.99         &-                \\

LayoutLMv3$_\mathrm{Large}$  \citep{huang2022layoutlmv3}       & 368M                  & T+L+I             & 92.08          & \textbf{97.46}         & -               \\
ERNIE-Layout$_\mathrm{Large}$  \citep{peng2022ernie}    & -                     & T+L+I             & 93.12          & 97.21         & 97.55          \\ \hline
\textbf{LayoutMask$_\mathrm{Large}$ (Ours)} & 404M                   & T+L               & \textbf{93.20$\pm$0.29}     & 97.19$\pm$0.20    & 97.27$\pm$0.32     \\ \hline
\end{tabular}
\caption{\label{table:sota}
F1 scores (\%) of different methods on FUNSD, CORD, and SROIE .The best results are denoted in boldface. $\dag$: TILT utilized supervised datasets during pre-training, so the scores are not directly comparable.
}
\end{table*}

\subsubsection{Masked Position Modeling}
\label{sec:mpm}

To further promote the representation learning of layout information in the MLM task, we design an auxiliary task, Masked Position Modeling (MPM), which has a symmetric pre-training objective: recovering randomly masked 2D positions during pre-training (See illustration in Figure \ref{fig:model-structure}).
Inspired by WWM, we also apply the MPM task at word-level instead of token-level.
For each pre-training document, we randomly choose some unduplicated words with a given probability $\mathrm{P}_\mathrm{mpm}$.
Then, for each selected word, we mask their 2D positions with the following two steps:

\noindent\textbf{Box Split:}
We first split the selected word out of its segment so the original segment box becomes 2 or 3 segment
pieces (depending on if the word is at the start/end or in the middle).
The selected word becomes a one-word segment piece with just itself.
Then we update the local 1D positions (restarting with 1) and segment 2D positions for each new segment piece.
With the above operations, we can eliminate the local reading order clues implied by original 1D and 2D positions, so the model has to focus on semantical clues and new 2D positions.

\noindent\textbf{Box Masking:}
For each selected word, we mask its 2D position with pseudo boxes: $[0,0,0,n]$ where $n\in [0,1,2,...]$ is a random number.
Notice that segment 2D position is shared among tokens in the same segment, so the pseudo boxes will act as identifiers to distinguish identical tokens from different masked boxes, thus avoiding ambiguity.  

During pre-training, our model is supposed to predict the masked 2D positions with GIoU loss \citep{rezatofighi2019generalized}:

\begin{equation}
\label{equ:mpm}
\small
\mathcal{L}_\mathrm{mpm}= 
- \frac{1}{N}
 \sum_{i=1}^{N} 
(\frac{|\mathrm{B}_{i} \cap \mathrm{\bar{B}}_{i}|}{|\mathrm{B}_{i} \cup \mathrm{\bar{B}}_{i}|}-
\frac{|\mathrm{C}_i \backslash(\mathrm{B}_{i} \cup \mathrm{\bar{B}}_{i})|}{|\mathrm{C}_i|}).
\end{equation}

Here, $i\in[1,2,...,N]$ is the index of $N$ masked 2D positions.
$\mathrm{B}_{i}$ is the ground truth box normalized to [0,1], and $\mathrm{\bar{B}}_{i}$ denotes the predicted 2D position.
$\mathrm{C}_i$ is the smallest convex shapes that covers $\mathrm{B}_{i}$ and $\mathrm{\bar{B}}_{i}$. $\mathcal{L}_\mathrm{mpm} $ is the average GIoU loss of $N$ masked 2D positions.

The MPM task is very similar to the cloze test, where a group of randomly selected words is supposed to be refilled at the right positions in the original document. 
To predict the masked 2D positions of selected words, the model has to find the context for each word based on semantic relations and then infer with 2D position clues from a spatial perspective. The joint learning process with both semantic and spatial inference can promote text-layout interactions and help the model to learn better layout representations.

With the above two pre-training objectives, the model is pre-trained with the following loss:
\begin{equation}
\label{equ:total}
\mathcal{L}_\mathrm{total}= \mathcal{L}_\mathrm{mlm}+\lambda\mathcal{L}_\mathrm{mpm},
\end{equation}
where $\lambda$ is a hyper-parameter that controls the balance of the two pre-training objectives.

\section{Experiments}

\subsection{Pre-training Settings}
LayoutMask is pre-trained with IIT-CDIP Test Collection \citep{IIT-CDIP-lewis2006building}.
It contains about 42 million scanned document pages, and we only use 10 million pages. 
We use a public OCR engine, PaddleOCR\footnote{https://github.com/PaddlePaddle/PaddleOCR} to obtain the OCR results.  

We train LayoutMask with two parameter sizes. LayoutMask$_\mathrm{Base}$ has 12 layers with 16 heads, and the hidden size is 768.
LayoutMask$_\mathrm{Large}$ has 24 layers with 16 heads where the hidden size is 1024.
LayoutMask$_\mathrm{Base}$ and LayoutMask$_\mathrm{Large}$ are initialized with pre-trained XLM-RoBERTa models \citep{conneau2020unsupervised}.

For hyper-parameters, we have $\mathrm{P}_\mathrm{mlm}$=25\% and $\mathrm{P}_\mathrm{mpm}$=15\% (See ablation study in Section \ref{appendix:masking-p} of the Appendix). The weight of MPM loss $\lambda$ is set to be 1.

\begin{table}[]
\centering
\scriptsize
\begin{tabular}{l|l|cc}
\hline
\multirow{2}{*}{\textbf{Method}} & \multirow{2}{*}{\textbf{Modality}} & \multicolumn{2}{c}{\textbf{Accuracy$\uparrow$}}       \\ \cline{3-4} 
                        &                           & \multicolumn{1}{c|}{$\mathrm{Base}$}  &$\mathrm{Large}$\\ \hline
VGG-16 \citep{afzal2017cutting}                 & I                         & \multicolumn{2}{c}{90.97}          \\
Ensemble    \citep{das2018document}           & I                         & \multicolumn{2}{c}{93.07}   \\ 
LadderNet   \citep{sarkhel2019deterministic}            & I                         & \multicolumn{2}{c}{92.77}          \\ \hline
BERT     \citep{kenton2019bert}                 & T                         & \multicolumn{1}{c|}{89.81} & 89.92 \\
RoBERTA      \citep{liu2019roberta}               & T                         & \multicolumn{1}{c|}{90.06} & 90.11 \\ \hline
UniLMv2      \citep{bao2020unilmv2}              & T+L                       & \multicolumn{1}{c|}{90.06} & 90.20 \\
LayoutLM  \citep{xu2020layoutlm}               & T+L                       & \multicolumn{1}{c|}{91.78} & 91.90 \\
StructuralLM     \citep{li2021structurallm}       & T+L                       & \multicolumn{1}{c|}{-}     & 96.08 \\ \hline
SelfDoc   \citep{li2021selfdoc}              & T+L+I                     & \multicolumn{1}{c|}{92.81} & -     \\
TITL       \citep{powalski2021going}                  & T+L+I                     & \multicolumn{1}{c|}{95.25} & 95.52 \\
LayoutLMv2   \citep{xu2021layoutlmv2}           & T+L+I                     & \multicolumn{1}{c|}{95.25} & 95.64 \\
DocFormer  \citep{appalaraju2021docformer}               & T+L+I                     & \multicolumn{1}{c|}{96.17} & 95.50 \\
LiLT      \citep{wang2022lilt}                 & T+L+I                     & \multicolumn{1}{c|}{95.68} & -     \\
LayoutLMv3 \citep{huang2022layoutlmv3}                & T+L+I                     & \multicolumn{1}{c|}{95.44} & 95.93 \\
ERNIE-Layout      \citep{peng2022ernie}       & T+L+I                     & \multicolumn{1}{c|}{-}     & 96.27 \\ \hline
\textbf{LayoutMask (Ours)}              & T+L                       & \multicolumn{1}{c|}{93.26} & 93.80 \\ \hline
\end{tabular}
\caption{\label{table:rvlcdip-sota}
The accuracies (\%) of different methods on RVL-CDIP dataset. For transformer-based models, we provide results for both base and large versions.}
\end{table}

\subsection{Comparison with the State-of-the-Art}
In this section, we compare LayoutMask with SOTA models on two VrDU tasks: form \& receipt understanding and document image classification.
\begin{table*}[!t]
\centering
\small
\begin{tabular}{cc|ccc}
\hline
\multicolumn{2}{c|}{\textbf{Position Settings}} & \multicolumn{3}{c}{\textbf{Datasets}}               \\ \hline
\textbf{1D}        & \textbf{2D}       & \textbf{FUNSD (F1$\uparrow$)}      & \textbf{CORD (F1$\uparrow$)}       & \textbf{SROIE (F1$\uparrow$)}      \\ \hline
Global                      & Word                       & 82.17$\pm$0.45          & 95.95$\pm$0.43          & 96.02$\pm$0.34          \\ \hline
Global                      & Segment                    & 91.61$\pm$0.42          & \textbf{96.69$\pm$0.24}          & 96.20$\pm$0.26          \\ \hline
Local                       & Word                       & 91.65$\pm$0.36          & 95.86$\pm$0.22          & 96.54$\pm$0.23          \\ \hline
Local                       & Segment                    & \textbf{92.30$\pm$0.24} & 96.68$\pm$0.12 & \textbf{96.56$\pm$0.21} \\ \hline
\end{tabular}
\caption{\label{ablation-on-position}
 The average F1 scores (\%) with different 1D position and 2D position combinations. The best results are denoted in boldface.
}
\end{table*}

\subsubsection{Form and Receipt Understanding}

In this task, we conduct entity extraction task on three document understanding datasets: FUNSD \citep{jaume2019funsd}, CORD \citep{park2019cord}, and SROIE \citep{huang2019icdar2019}.
The FUNSD dataset is a form understanding dataset, which contains 199 documents (149 for training and 50 for test) and 9707 semantic entities.
The CORD dataset is a receipt understanding dataset, and it contains 1000 receipts (800 for training, 100 for validation, and 100 for test) with 30 semantic labels in 4 categories.
The SROIE dataset is another receipt understanding dataset with four types of entities, containing 626 receipts for training and 347 receipts for test.

For evaluation, we adopt the word-level F1 score as the evaluation metric for FUNSD and CORD and use the entity-level F1 score for SROIE.
Since these datasets are quite small, in order to provide stable and reliable results, we repeat our experiments ten times for each test and report the average F1 scores and standard errors as the final results.

The results of previous methods and LayoutMask on these datasets are listed in Table \ref{table:sota}.
We have categorized them by the modalities used in pre-training: ``T'' for text, ``L'' for layout, and ``I'' for image. 
Notice that LayoutMask is a ``T+L'' model that does not use image modality. 

For the base version, LayoutMask$_\mathrm{Base}$ outperforms other methods, including ``T+L+I'' models, on all three datasets (FUNSD+2.62\%, CORD+0.43\%, SROIE+0.62\%).
For the large version, LayoutMask$_\mathrm{Large}$ ranks first on FUNSD and has comparable results on CORD and SROIE.

These results show that LayoutMask has competitive performance with SOTA methods, demonstrating the effectiveness of our proposed modules.
Since LayoutMask only uses text and layout information, we believe that the potential power of layout information has not been fully explored in previous studies. 

\subsubsection{Document Image Classification}
In the document image classification task, we aim to classify document images in RVL-CDIP dataset \citep{rvl-cdip-harley2015evaluation}.
This dataset is a subset of the IIT-CDIP collection with 400,000 labeled document images (320,000 for train, 40,000 for validation, and 40,000 for test) in 16 categories. We use PaddleOCR to extract text and layout information as model input. 
We compare different methods with the overall classification accuracies on RVL-CDIP, and the results are in Table \ref{table:rvlcdip-sota}. 

It is observed that LayoutMask has beaten all uni-modality models (``I'' and ``T'').
For ``T+L'' models, LayoutMask$_\mathrm{Base}$ outperforms other base models with a margin of 1.48\%, while LayoutMask$_\mathrm{Large}$ takes the second place in large models. 
Compared with ``T+L+I'' models where image modality is utilized, LayoutMask falls behind due to the lack of visual features from image modality. 
We have found that the image modality plays an important role in this task because RVL-CDIP images contain many elements that cannot be recognized by OCR engines (\eg figures, table lines, and handwritten texts) and have orientation issues (See examples in Figure \ref{fig:rvl-cdip} of the Appendix). 
So the lack of image modality will bring difficulties that cannot be solved with only text and layout information.

\subsection{Ablation Study on LayoutMask}

\begin{table*}[ht]
\centering
\small
\begin{tabular}{c|c|c|ccccc}
\hline
\multirow{2}{*}{\textbf{\#}}           &\multirow{2}{*}{\textbf{\makecell[c]{Position Settings\\1D \& 2D}}}           & \multirow{2}{*}{\textbf{\makecell[c]{Swap\\ Probability}}}          & \multicolumn{5}{c}{\textbf{SROIE (F1$\uparrow$)}}                                                                                   \\ \cline{4-8}
    &      & & \textbf{Address} & \textbf{Company} & \textbf{Date} & \multicolumn{1}{c|}{\textbf{Total}} & \multicolumn{1}{c}{\textbf{Overall}} \\ \hline	 	 	 
1 &Local+Segment                   & -           & 96.69$\pm$0.37            & 95.88$\pm$0.28             & 99.66$\pm$0.13         & \multicolumn{1}{c|}{94.02$\pm$0.49}          & \multicolumn{1}{c}{96.56$\pm$0.21}            \\ \hline 	 	 	 	 
2 &\multirow{4}{*}{Global+Segment} & -                       & 96.54$\pm$0.51            & 95.84$\pm$0.59            & 99.69$\pm$0.26         & \multicolumn{1}{c|}{92.73$\pm$0.57}          & \multicolumn{1}{c}{96.20$\pm$0.26}            \\
 	 	 	 	 
 3&                                & 10                       & 91.73$\pm$2.00            & 95.22$\pm$0.61            & 99.65$\pm$0.34         & \multicolumn{1}{c|}{91.87$\pm$1.33}          & \multicolumn{1}{c}{94.62$\pm$0.69}            \\
 4 &                               & 20                       & 90.03$\pm$3.77            & 94.93$\pm$0.53            & 99.60$\pm$0.32         & \multicolumn{1}{c|}{91.67$\pm$1.35}          & \multicolumn{1}{c}{94.06$\pm$1.02}            \\
 	 	 	 	 
 5&                               & 30                       & 88.12$\pm$4.59            & 94.88$\pm$0.82            & 99.55$\pm$0.28         & \multicolumn{1}{c|}{91.19$\pm$1.38}          & \multicolumn{1}{c}{93.44$\pm$1.14}            \\ \hline
\end{tabular}
\caption{\label{table-robutsness}
F1 scores (\%) on SROIE dataset with difference 1D positions and increasing segment swap probabilities (\%). We report both entity-level scores (``Address'', ``Company'', ``Date'', and ``Total'') and overall results (``Overall'') for detailed comparison.
}
\end{table*}

\subsubsection{Comparison of Layout Information }
\label{sec:layout-ablation}
We first compare the performance of LayoutMask using different layout information.
To make a fair comparison, we use LayoutMask with only the MLM task and the WWM strategy during pre-training.
For each test, LayoutMask is pre-trained and fine-tuned with a specific 1D and 2D position combination.
The results are listed in Table \ref{ablation-on-position}. 

\noindent\textbf{Performance of 1D Position:}
For 1D position, Local-1D outperforms Global-1D on both FUNSD (+9.48\%/+0.69\% with Word-2D/Segment-2D) and SROIE (+0.52\%/+0.36\%) and falls a little behind on CORD (-0.07\%/-0.01\%). 

\begin{figure}[t]
	\centering
	\includegraphics[width=7.5cm]{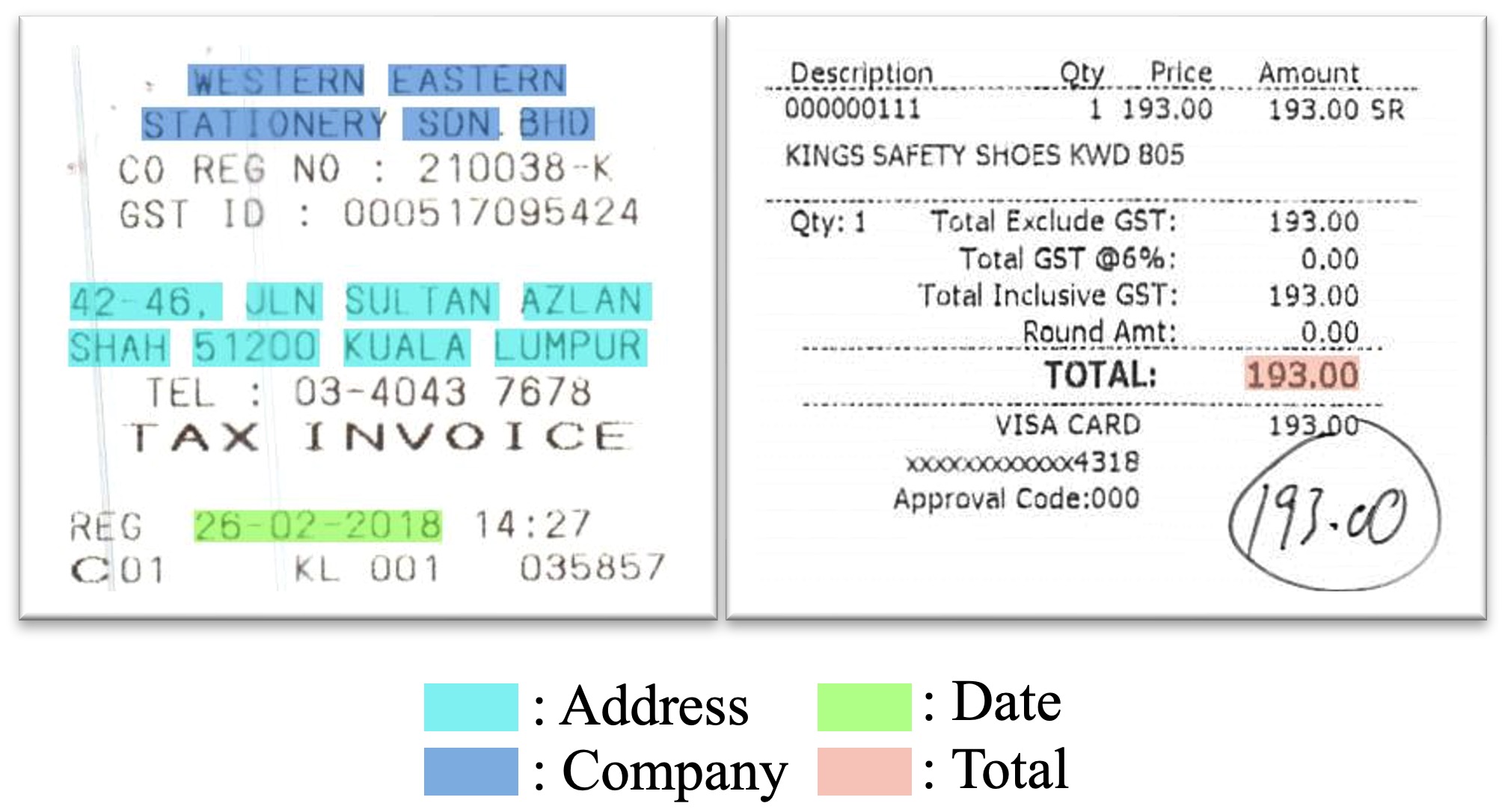}
	\caption{Two images from SROIE dataset. Colored boxes denote the ground truth of entities. The left image contains two cross-line and cross-segment entities (``Address'' and ``Company''). The right image, with a mixture of vertical and horizontal layouts, contains the ``Total'' entity.}
	\label{fig:sroie-error}
\end{figure}

To understand the benefits of using Local-1D, we provide entity-level F1 score on SROIE dataset in Table \ref{table-robutsness} (\#1 for Local+Segment and \#2 for Global+Segment). 
It is obvious that the performance gap between Local+Segment and Global+Segment mainly comes from entity ``Total'' (from 94.02\% to 92.72\%), while other entities have similar F1 scores.
We illustrate two example images of SROIE and their entities annotations in Figure \ref{fig:sroie-error}.
The right image, which contains entity ``Total'', has both vertical layout (first two lines) and horizontal layout and has multiple misleading numbers with the same content as ground truth (\ie ``193.00'').
So it is hard to recognize the entity ``Total'' by using the ordinary reading order implied by Global-1D. 
Therefore, using Local-1D can perform better since it is more adaptive to such cases.

\noindent\textbf{Performance of 2D Position:}
For 2D position, using segment-level 2D position brings better results on all three datasets, regardless of the 1D position types. 
An important reason is that the segment information is highly indicative of recognizing entities.
For example, every entity in FUNSD and CORD exactly shares the same segment. 
Therefore, although Word-2D contains more layout details, it will break the alignments between 2D positions and entities, thus bringing performance drops.
A typical result of such phenomenon\footnote{Similar phenomenon can also be observed in the LayoutLM series models, where using Segment-2D increase F1 scores for about 8\% on FUNSD dataset.} can be seen on FUNSD, where replacing Global+Segment to Global+Word will result in a significant decrease of 9.44\%.

\begin{table*}[]
\centering
\small
\begin{tabular}{c|cccc|cccc}
\hline
\multirow{2}{*}{\textbf{\#}} & \multicolumn{4}{c|}{\textbf{Pre-training Setting}}        & \multicolumn{4}{c}{\textbf{Datasets}}                                 \\ \cline{2-9} 
                             & \textbf{MLM} & \textbf{WWM} & \textbf{LAM} & \textbf{MPM} & \textbf{FUNSD (F1$\uparrow$)} & \textbf{CORD (F1$\uparrow$)} & \textbf{SROIE (F1$\uparrow$)}  & \textbf{RVL-CDIP (ACC$\uparrow$)} \\ \hline
1                            &       $\surd$       &              &              &              & 89.73$\pm$0.50       & 96.32$\pm$0.15    & 95.76$\pm$0.34     & 92.17              \\
2                            &         $\surd$      &        $\surd$       &              &              & 92.30$\pm$0.24      & 96.68$\pm$0.12    & 96.56$\pm$0.21     & 92.89              \\
3                            &          $\surd$     &    $\surd$           &     $\surd$          &              & 92.66$\pm$0.26      & 96.89$\pm$0.24    & 96.64$\pm$0.22     & 93.03              \\
4                            &       $\surd$        &          $\surd$     &          &       $\surd$        & 92.77$\pm$0.30      & 96.84$\pm$0.17    & 96.66$\pm$0.32     & 93.11              \\ 
5                            &       $\surd$        &          $\surd$     &     $\surd$          &       $\surd$        & 92.91$\pm$0.34      & 96.99$\pm$0.30    & 96.87$\pm$0.19     & 93.26              \\ \hline
\end{tabular}
\caption{\label{table:ablation-on-tasks}
Performance analysis with different pre-training objectives and masking strategies.
}
\end{table*}

\noindent\textbf{Robustness Comparison:}
Besides performance superiority, another important reason to choose the local 1D position is its robustness to layout disturbance.
In real-world cases, a typical layout disturbance is ``Segment Swap'', where segments in the same line are indexed with wrong orders due to document rotation or OCR issues.
In such scenarios, the incorrect cross-segment order will lead to incorrect global 1D positions and can be harmful to model inference. Fortunately, the local 1D position is naturally immune to such disturbance since it does not rely on cross-segment orders, making it more robust than global 1D position.

To quantify such differences in robustness, we demonstrate how the segment swap will influence the performance of using global 1D position by simulating it on test datasets.
For each test document, we randomly choose some lines with a given probability $\mathrm{P}_\mathrm{swap}$ and then swap the segments in it. 
We conduct experiments on LayoutMask$_\mathrm{Base}$ (MLM+WWM) in Global+Segment setting with different $\mathrm{P}_\mathrm{swap}$ (\ie 10\%, 20\%, and 30\%) and the results are reported in Table \ref{table-robutsness} (\#3-5).

During our experiments, we have found that the segment swap does not bring significant performance changes on FUNSD and CORD datasets (so these results are not listed due to the limited space). 
A possible reason is that FUNSD and CORD do not contain cross-segment entities, so the segment swap can not break the order of words in each entity. 
Evidence for this explanation is that the SROIE dataset is significantly affected by segment swap, and its cross-segment entities (``Address'' and ``Company'') have obvious performance drops.
In SROIE, the majority of ``Address'' entities and a few ``Company'' entities are printed in multiple lines (See examples in Figure \ref{fig:sroie-error}), so the segment swap can change the in-entity orders of entity words. The results show that the ``Address'' entity has the largest drop among all entities (-4.81\%, -6.51\%, and -8.42\% for $\mathrm{P}_\mathrm{swap}$=10\%, 20\%, 30\%).
Besides, the ``Total'' entity has the second largest decrease (-0.86\%, -1.06\%, and -1.54\%). 
As aforementioned, the ``Total'' entities are usually surrounded by complex layouts and misleading numbers, so the segment swap will bring extra difficulties in recognizing the correct entities. 

The above performance decreases of using global 1D position prove the superiority of using local 1D position since it is not affected by such layout disturbance and can have more robust performance in real-world scenarios.

\subsubsection{Effectiveness of Proposed Methods}


In Table \ref{table:ablation-on-tasks}, we provide results using different pre-training tasks and masking strategies to demonstrate the effectiveness of our proposed modules. 

Comparing \#1 and \#2 in Table \ref{table:ablation-on-tasks}, we observe that WWM brings significant performance improvements on all datasets. The reason is that it increases the difficulty of the MLM task, so we can obtain a stronger language model.  
We also find that LAM can also brings consistent improvements on all dataset because LAM can force the model to learn better representations for layout information, which is beneficial to downstream tasks. 

Comparing \#2 to \#4 and \#3 to \#5, it is observed that the MPM task also brings considerable improvements on all datasets. MPM works as an auxiliary task to help the MLM task and can increase the pre-training difficulty, contributing to learning better and more robust layout representations.

Moreover, the full-version LayoutMask  (\#5) outperforms the naive version (\#1) by a large margin (FUNSD+3.18\%, CORD+0.67\%, SROIE+1.11\%, and RVL-CDIP+1.09\%), demonstrating the effectiveness of our proposed modules when working together.
To better illustrate the effectiveness of our model design, we list category-level accuracy improvements on RVL-CDIP dataset and provide detailed discussions in Section \ref{appendix:rvl-cdip} of the Appendix.

\section{Conclusion}
In this paper, we propose LayoutMask, a novel multi-modal pre-training model, to solve the reading order issues in VrDU tasks.
LayoutMask adopts local 1D position as layout input and can generate adaptive and robust multi-modal representations.
In LayoutMask, we equip the MLM task with two masking strategies and design a novel pre-training objective, Masked Position Modeling, to enhance the text-layout interactions and layout representation learning. 
With only using text and layout modalities, our method can achieve excellent results and significantly outperforms many SOTA methods in VrDU tasks. 

\section*{Limitations}

Our method has the following limitations:

\noindent\textbf{Datasets:}
In multi-modal pre-training, we rely on downstream datasets to evaluate the performance of pre-trained models.
The commonly used entity extraction datasets are relatively small and lack diversity, so the proposed method may not generalize well to real word scenarios.

\noindent\textbf{Lack of Image Modality:}
In LayoutMask, we focus on text-layout interactions, leaving the image modality unexplored.
However, documents in the real world contain many elements that can not be described by text and layout modalities, like figures and lines, so incorporating image modality is important in building a universal multi-modal pre-training model for document understanding.

\bibliography{anthology,custom}
\bibliographystyle{acl_natbib}

\newpage
\appendix

\section{Ablation Study of Masking Probabilities}
\label{appendix:masking-p}
\begin{figure}[!tb]
	\centering
	\includegraphics[width=7cm]{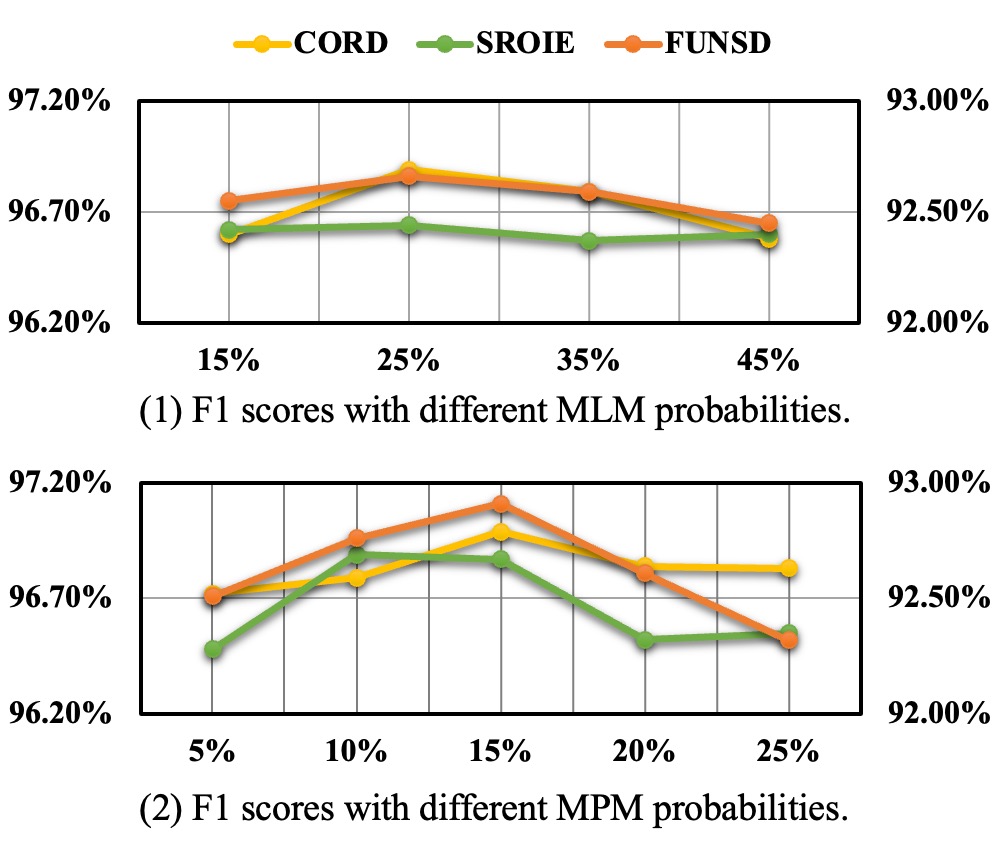}
	\caption{The F1 scores on FUNSD, CORD, and SROIE with different masking probabilities. FUNSD dataset uses the x-axis on the right side.}
	\label{fig:probability}
\end{figure}

We compare LayoutMask using different $\mathrm{P}_\mathrm{mlm}$ and $\mathrm{P}_\mathrm{mpm}$, and the results are in Figure \ref{fig:probability}.
We first find the best $\mathrm{P}_\mathrm{mlm}$ without using the MPM task, and the optimal value is 25\%. 
Then we fix such optimal $\mathrm{P}_\mathrm{mlm}$ to find the best $\mathrm{P}_\mathrm{mpm}$, which is 15\% as the results show.

\label{sec:ablation-method}

\begin{figure}[!tb]
	\centering
	\includegraphics[width=7.5cm]{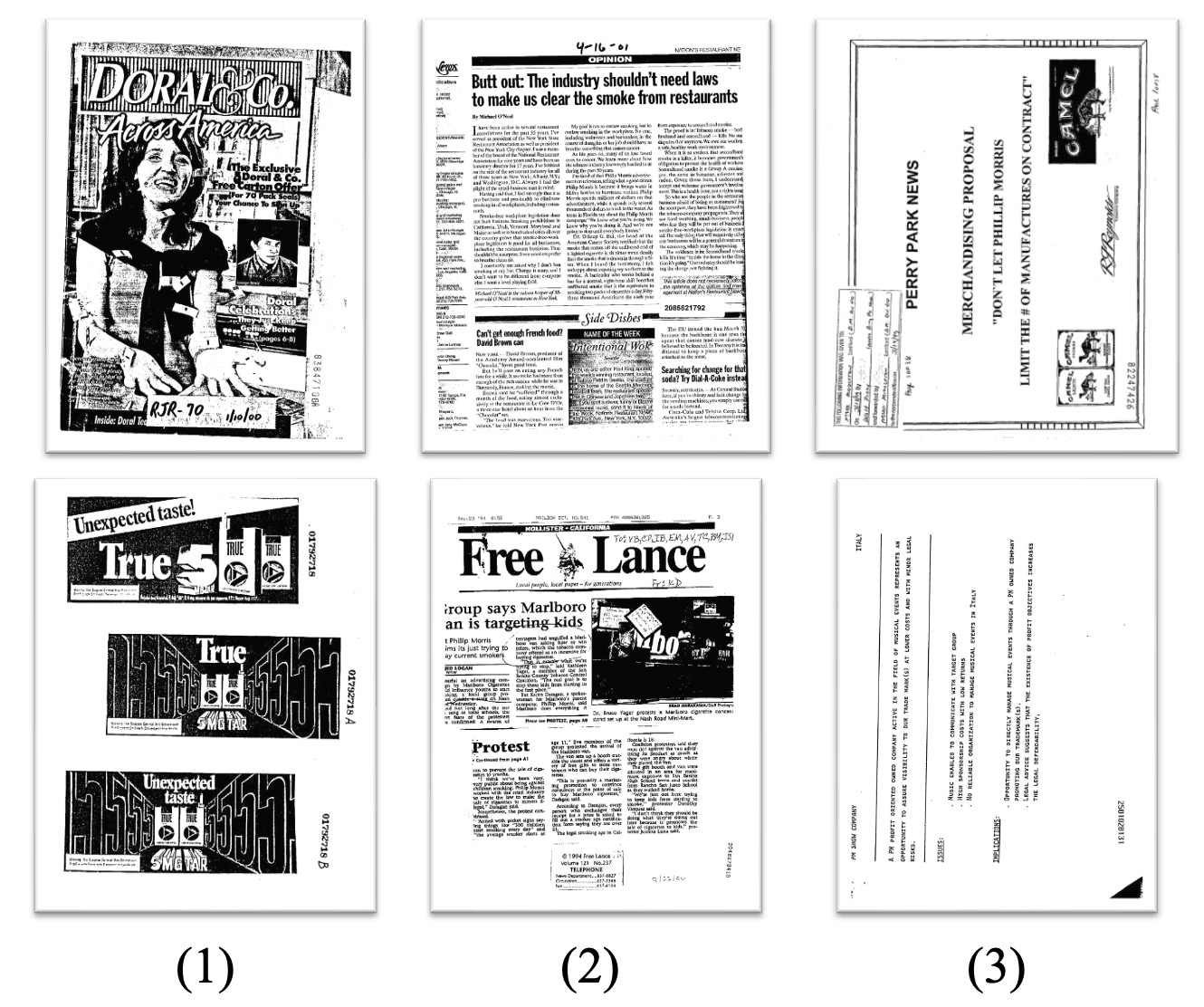}
	\caption{ RVL-CDIP images from different categories. (1) Advertisement; (2) News article; (3) Presentation (with incorrect orientations).}
	\label{fig:rvl-cdip}
\end{figure}

\section{Ablation Study on RVL-CDIP}
\label{appendix:rvl-cdip}
To further understand the effectiveness of our model design, we list the detailed classification results on RVL-CDIP dataset with the naive version and the full version in Table \ref{table:rvlcdip-compare}.  
It is observed that the major performance improvements come from three categories: presentation (+3.36\%), advertisement (+2.93\%), and news article (+2.35\%).
We find these categories have more diverse layouts (See examples in Figure \ref{fig:rvl-cdip}), so classifying these documents requires a better understanding of the document structure, which also indicates the effectiveness of our methods in helping layout understanding.

 \begin{table}[!tb]
\centering
\begin{tabular}{l|cc|c}
\hline
\multirow{2}{*}{\textbf{Category}} & \multicolumn{2}{c|}{\textbf{Model Settings}} & \multirow{2}{*}{\textbf{Diff. (\%)}} \\ \cline{2-3}
        & \textbf{\textit{Naive}} & \textbf{\textit{Full}} & \\ \hline
letter         & 90.30      & 90.86   & 0.56 \\
form          & 85.71      & 86.77   & 1.07 \\
email         & 98.17      & 98.33   & 0.15 \\
handwritten      & 93.96      & 94.26   & 0.30 \\
\textbf{advertisement}     & 88.47      & 91.40   & \textbf{2.93} \\
sci-report   & 87.87      & 89.38   & 1.51 \\
sci-publication & 93.08      & 93.73   & 0.65 \\
specification     & 95.91      & 96.56   & 0.64 \\
file folder      & 91.29      & 92.71   & 1.42 \\
\textbf{news article}      & 90.09      & 92.44   & \textbf{2.35} \\
budget         & 94.01      & 94.96   & 0.95 \\
invoice        & 94.02      & 94.54   & 0.52 \\
\textbf{presentation}      & 86.14      &89.50   & \textbf{3.36} \\
questionnaire     & 92.44      & 92.88   & 0.44 \\
resume         & 98.31      & 98.70   & 0.39 \\
memo          & 94.93      & 95.12   & 0.19 \\ \hline
Overall        & 92.17      & 93.26   & 1.09 \\ \hline
\end{tabular}
\caption{\label{table:rvlcdip-compare}
The category-level accuracies (\%) on RVL-CDIP dataset of LayoutMask on the naive version and the full version. Categories with top-3 accuracy improvements are denoted in boldface.
}
\end{table}

\end{document}